\pdfoutput=1

\documentclass[11pt]{article}

\usepackage{acl}

\usepackage{lipsum}

\usepackage{times}
\usepackage{latexsym}

\usepackage[T1]{fontenc}

\usepackage[utf8]{inputenc}

\usepackage{microtype}

\usepackage{xcolor}
\usepackage{colortbl}
\usepackage{amssymb}
\usepackage{amsmath}
\usepackage{soul}
\usepackage{algorithm}
\usepackage{algorithmic}
\usepackage{booktabs}
\usepackage{multirow}
\usepackage{array}
\usepackage{graphicx}
\usepackage{caption}
\usepackage{subcaption}
\usepackage{enumitem}
\usepackage{tabularx}
\usepackage{float}

\definecolor{green}{RGB}{53, 130, 74}
\definecolor{orange}{RGB}{209, 154, 0}
\definecolor{lightgray}{RGB}{239, 239, 239}
\usepackage{xspace}

\title{StereoKG: Data-Driven Knowledge Graph Construction \\ for Cultural Knowledge and Stereotypes}

\author{Awantee Deshpande, Dana Ruiter, Marius Mosbach, Dietrich Klakow \\
  Spoken Language Systems Group, Saarland University, Germany \\
  \texttt{\{adeshpande,druiter,mmosbach,dietrich\}@lsv.uni-saarland.de}}

\begin{document}
\maketitle

\begin{abstract}
Analyzing ethnic or religious bias is important for improving fairness, accountability, and transparency of natural language processing models. However, many techniques rely on human-compiled lists of bias terms, which are expensive to create and are limited in coverage. In this study, we present a fully data-driven pipeline for generating a knowledge graph (KG) of cultural knowledge and stereotypes. Our resulting KG covers 5 religious groups and 5 nationalities and can easily be extended to include more entities. Our human evaluation shows that the majority (59.2\%) of non-singleton entries are coherent and complete stereotypes. We further show that performing intermediate masked language model training on the verbalized KG leads to a higher level of cultural awareness in the model and
has the potential to increase classification performance on knowledge-crucial samples on a related task, i.e., hate speech detection.
\end{abstract}

\section{Introduction}

Fairness, accountability, and transparency to fight model-inherent bias and discrimination have become a major branch of machine learning research in recent years. This includes studying cultural bias and stereotypes in datasets and language models. Stereotypes are cognitive schemas that aid in categorizing and perceiving other social groups \citep{hilton1996stereotypes}, and becoming conscious of this stereotyping can increase cultural knowledge and sensitivity \citep{buchtel2014cultural}. However, without mindfulness, stereotypes lead to inferring traits of individuals from their (e.g., socio-economic) status or social group \citep{hoffman1990gender}, which then leads to systemic discrimination. Stereotypes as inherent cognitive functions are equally present in human-generated content, e.g., text resources used to train machine learning algorithms, which then further propagate and lead to discrimination \citep{hovy-spruit-2016-social}. 
Within the natural language processing community, bias reduction includes work in reducing gender \citep{bolukbasi2016debiasing}, ethnic or religious bias \citep{manzini-etal-2019-black} in word embeddings or classification tasks \citep{dixon2018measuring,badjatiya2020stereotypical, mozafari2020hate}. Nevertheless, these techniques often rely on predefined lexicons,
which are mostly human-written and thus expensive in their creation.
Instead, we present an entirely data-driven pipeline for the creation of a scalable knowledge graph (KG) of cultural knowledge and stereotypes. Our resulting knowledge graph, called StereoKG, consists of 4,722 entries about 10 different social groups, i.e., 5 religious groups and 5 nationalities. This knowledge graph has several use cases, ranging from analyzing existing stereotypical and cultural knowledge online, to removing ethnic and religious bias or increasing the cultural awareness of classifiers. In our experiments, we focus on the latter: integration of cultural knowledge to improve classification performance.
Overall, our contributions are threefold:

\begin{itemize}
    \item Development of a fully \textbf{data-driven} knowledge graph construction approach on Twitter and Reddit data.
    \item \textbf{Manual evaluation} and analysis of the resulting knowledge graph of cultural knowledge and stereotypes, highlighting the importance of multiple-mention entries in representing cultural stereotypes, which achieve higher quality than single-mention entries.
    \item Classification experiments showing that performing intermediate masked language model training on linearized stereotype knowledge can improve the \textbf{classification performance} on knowledge-crucial samples on a hate speech task.
\end{itemize}

The rest of the paper is structured as follows:  After describing the related work (Section \ref{s:rel_work}), we present our knowledge graph creation technique (Section \ref{s:kg_creation}) which is then evaluated in a quantitative and qualitative fashion (Section \ref{s:kg_eval}).
Section \ref{s:knowledge_integration} describes the knowledge integration experiments, which constitute downstream task performance on hate speech detection
and masked language modelling predictions of cultural content. We then discuss (Section \ref{s:discussion}) and conclude (Section \ref{s:conclusion}) our findings. Ethical concerns are addressed in Appendix \ref{a:ethics}.

\section{Related Work}
\label{s:rel_work}

\textbf{Cultural knowledge} about different social groups and entities plays an important role in responding to contextual situations. In this work, we target cultural knowledge as a form of commonsense \citep{lobue-yates-2011-types-commonsense}.
Incorporating cultural commonsense in reasoning tasks is an understudied practice in NLP.
\citet{anacletoCanCS} 
study the variation of cultural commonsense and how it affects computer applications. 
While there exist knowledge base resources for general commonsense \cite{cyc-10.1145/219717.219745, conceptnet-DBLP:journals/corr/SpeerCH16, webchild-10.1145/2556195.2556245}, to the best of our knowledge,
\citet{acharya-atlas2020} have provided the only work targeting the construction of a cultural knowledge graph, which comprises various rituals and customs for two cultures.
However, since it relies on crowdsourcing, it is limited in its coverage and is not easily extendable.

Cultural knowledge is largely correlated to \textbf{stereotypes}. 
Contrary to exhaustive research avenues analyzing gender 
and ethnic stereotypes, our work focuses on the lesser-studied nationality and religious stereotypes. \citet{Snefjella-nationalstereo} have shown that national stereotypes could be grounded in the collective linguistic behavior of nations, while the Harvard Pluralism Project\footnote{\url{https://pluralism.org/stereotypes-and-prejudice}} stresses the importance of considering religion as a factor for prejudice. Because of the diversity of social groups and their behavioral traits, stereotypes and cultural attributes have 
unclear boundaries, making it difficult to distinguish between the two. Keeping this in mind, we treat cultural knowledge and stereotypes as interchangeable terms.

Stereotypes have been used to estimate \textbf{bias} in language models using curated datasets \cite{nadeem:stereoset, nangia:CrowPairs}. 
Stereotypical data has also been extracted from search engine autocomplete predictions using query prompts \citep{baker-google-stereotypes} and then used for analyzing how language models learn these concepts \cite{choenni2021}. \citet{bolukbasi2016debiasing} use minimal pairs of male-female terms to debias word embeddings.

In our work, we create a unified resource of cultural knowledge and stereotypes. Knowledge graphs serve as sources of representing knowledge in a structured format. Factual knowledge bases such as DBPedia \cite{dbpedia}, Freebase \cite{Bollacker2008:FreebaseAC} and Wikidata \cite{wikidata} contain grounded knowledge about individual entities. \textbf{Knowledge graph construction} for commonsense reasoning has also been a common object of research \citep{cyc-10.1145/219717.219745,conceptnet-DBLP:journals/corr/SpeerCH16,webchild-10.1145/2556195.2556245}.
While some KGs comprise an if-then reasoning scheme \citep{sapAtomic, forbes-choi-2017-verb}, some contain knowledge in the form of triples \citep{wikidata} or as simple natural language sentences \citep{GenericsKB-https://doi.org/10.48550/arxiv.2005.00660, neuralDB-https://doi.org/10.48550/arxiv.2010.06973}. Crowdsourced KGs, e.g., Wikidata, result in good quality knowledge, but require large-scale manual annotation and resources. In contrast, KGs constructed in an automated manner have a lower cost in construction, are easily extendable, and have been shown to be useful in several downstream applications \cite{yago-10.1145/1242572.1242667, GenericsKB-https://doi.org/10.48550/arxiv.2005.00660}. For example, \citet{romero2019_quasimodo} use questions as prompts for learning commonsense cues from search engine query logs and question-answering forums and construct a commonsense knowledge base.

Explicit \textbf{knowledge integration} of knowledge resources into language models can be roughly categorized into fusion based approaches and language modeling based approaches. Fusion based approaches \citep{peters-etal-2019-knowledge, KAdapter-DBLP:journals/corr/abs-2002-01808, K-XLNet-DBLP} typically perform knowledge integration by combining language model representations with representations extracted from knowledge bases. Compared to language modeling based approaches, as explored by us, they rely on aligned data and are typically applied during the pre-training stage.
Language modeling based approaches commonly start from a pre-trained language model and perform knowledge integration via intermediate pre-training. For example, \citet{bosselut-etal-2019-comet} integrate commonsense knowledge by performing language modeling on triples obtained from ATOMIC and ConceptNet. Recently, \citet{da2021analyzing} analyzed this approach in the few-shot training setting. In contrast to our study, both works consider autoregressive language models and use the resulting models for knowledge base construction, while we study the impact of knowledge integration on downstream task performance. Similar to our work, \citet{lauscher-etal-2020-common} integrate commonsense knowledge via masked language modeling. They obtain sentences for intermediate pre-training by randomly traversing the ConceptNet knowledge graph. Unlike our work, they do not update the weights of the pre-trained model and train adapter layers instead. Moreover, while we focus on hate-speech classification as our downstream task, they evaluate on GLUE.

\begin{figure*}[!t]
    \centering
    \includegraphics[width=\linewidth]{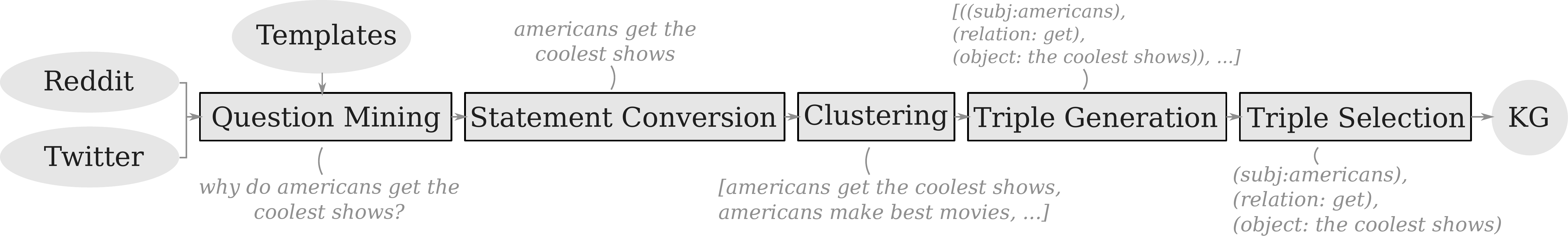}
    \caption{From noisy social media content to structured knowledge graph: the creation pipeline of StereoKG.}
    \label{f:pipeline}
\end{figure*}

\section{Knowledge Graph Construction}
\label{s:kg_creation}

We focus our cultural KG on 5 religious (\textit{Atheism\footnote{Although atheism is not a religion, we still include it under the list of religious dispositions as a religious belief.}, Christianity, Hinduism, Islam, Judaism}) and 5 national (\textit{American, Chinese, French, German, Indian}) entities.
Previous work on automatic KG creation depended on external algorithms, i.e., autocompletion of search engine queries \cite{romero2019_quasimodo, choenni2021,baker-google-stereotypes}. This dependency is limiting, as external providers may filter\footnote{In its battle against biased or hateful content, Google has imposed filters on its autocomplete predictions for targeted questions.} outputs of their autocomplete algorithm, especially on sensitive topics such as \emph{culture} and \emph{identity}. Instead, we keep control over the whole KG creation process. The entire KG construction pipeline is illustrated in Figure \ref{f:pipeline}.

Using statement and \textbf{question mining}, cultural knowledge and stereotypes regarding our entities of interest are collected from two social media platforms, Reddit and Twitter. For Reddit, we limit our search to subreddits relevant for the respective subjects (e.g. \emph{r/germany} for Germans) together with common question-answering subreddits (e.g., \emph{r/AskReddit}) using the PRAW\footnote{\url{https://github.com/praw-dev/praw}} library. The complete list of queried subreddits is given in Appendix \ref{a:subreddits}. Similar to the commonsense mining approach by \citet{romero2019_quasimodo} and \citet{choenni2021}, we use fixed question and statement templates (Table \ref{t:query_temp}) to identify potential sentences containing cultural knowledge with the assumption that questions posted about various national and religious entities act as cues for underlying stereotypical notions about them. This results in 11,259 mined questions and statements. The questions are then \textbf{converted into statements} using \texttt{Quasimodo}\footnote{\url{https://github.com/Aunsiels/CSK}} \cite{romero2019_quasimodo}, as OpenIE does not process interrogative sentences.

\begin{table}[t]
\centering
\small
\begin{tabular}{c} 
 \toprule

 \textbf{Query Templates}  \\ [0.5ex] 
 \midrule
 Why is \emph{<SUB>} \\
 Why isn't \emph{<SUB>} \\
 Why are \emph{<SUB>} \\
 Why aren't \emph{<SUB>} \\
 Why can \emph{<SUB>} \\
 Why can't \emph{<SUB>} \\
 Why do \emph{<SUB>} \\
 Why don't \emph{<SUB>} \\
 Why doesn't \emph{<SUB>} \\
 How is \emph{<SUB>} \\
 How do \emph{<SUB>} \\
 What makes \emph{<SUB>} \\
 Why does \emph{<SUB>} culture \\
 \midrule
 \emph{<SUB>} are so \\
 \emph{<SUB>} is such a \\
 \bottomrule
\end{tabular}
\caption{
Question-based (top) and statement-based (bottom) query templates.
}
\label{t:query_temp}
\end{table}

To reduce redundancies in the KG triples, we \textbf{cluster} the mined sentences with similar content together using the fast clustering method for community detection implemented in the \texttt{SentenceTransformers}\footnote{\url{https://www.sbert.net/examples/applications/clustering/README.html}} \citep{sbert:dblp} library. 
This step results in 6,993 singletons and 610 clusters with more than one instance. We hypothesize that non-singleton clusters are better representatives of cultural knowledge and stereotypes, as these are based on questions that have been asked by several users, while singletons may be based on unique thoughts which do not represent a popular stereotype or cultural reality. The qualitative difference between singletons and clusters is evaluated in Section \ref{ss:human_evaluation}.

All assertions are then \textbf{converted into triples} using \texttt{OpenIE} \cite{openie-mausam-2016}. As OpenIE outputs multiple triples which may be noisy or irrelevant, they are filtered using the following heuristics:

\begin{itemize}[topsep=0.5pt,itemsep=-1ex,partopsep=1ex,parsep=1ex]
    \item Eliminate triples containing personal pronouns, e.g., \textit{I}, \textit{he}.
    \item Eliminate triples not containing the original subject entity.
    \item Remove colloquialisms (e.g, \textit{lol}) and modalities (e.g., \textit{really}) from triples.
\end{itemize}

While most triples are singletons, many are part of a cluster. In order to \textbf{select the triple} to represent a cluster in the final KG, triples within a cluster are converted into sentences via concatenation of their subject-predicate-object terms. These are ranked on their grammaticality using a binary classification model\footnote{\url{https://huggingface.co/textattack/distilbert-base-uncased-CoLA}} trained on the corpus of linguistic acceptability (CoLA) \cite{cola:warstadt2018neural}. Concretely, the rank of a sentence is the score assigned to the \textit{grammatical} class by the classification model, and the triple with the highest rank is chosen as the representative for the entire cluster.
Since CoLA and the resulting classifier are restricted to English, our triple selection currently only works for English data.  However, our method provides an advantage over standard cluster representative selection methods such as centroid identification, since we ensure that the chosen representative triple is the most fluent choice in its cluster. This is important, since (grammatical) completeness is an important quality feature for a KG, which we also assess as part of our human evaluation.

\section{Knowledge Graph Evaluation}
\label{s:kg_eval}
The resulting KG consists of 4,722 entries, with Americans being the largest represented group (1,071 entries) and Jews (43) the smallest. The proposed pipeline can also be utilised to extend the KG with additional entities. In the following section, we describe the qualitative and quantitative evaluation of the generated KG.

\subsection{KG Statistics}
\label{ss:kg_stats}
To gain insights into the sentiments and overall distribution of descriptive predicates, we evaluate the KG on two criteria.

\paragraph{Sentiment Analysis}

\begin{figure}[th]
    \centering
    \includegraphics[width=\columnwidth]{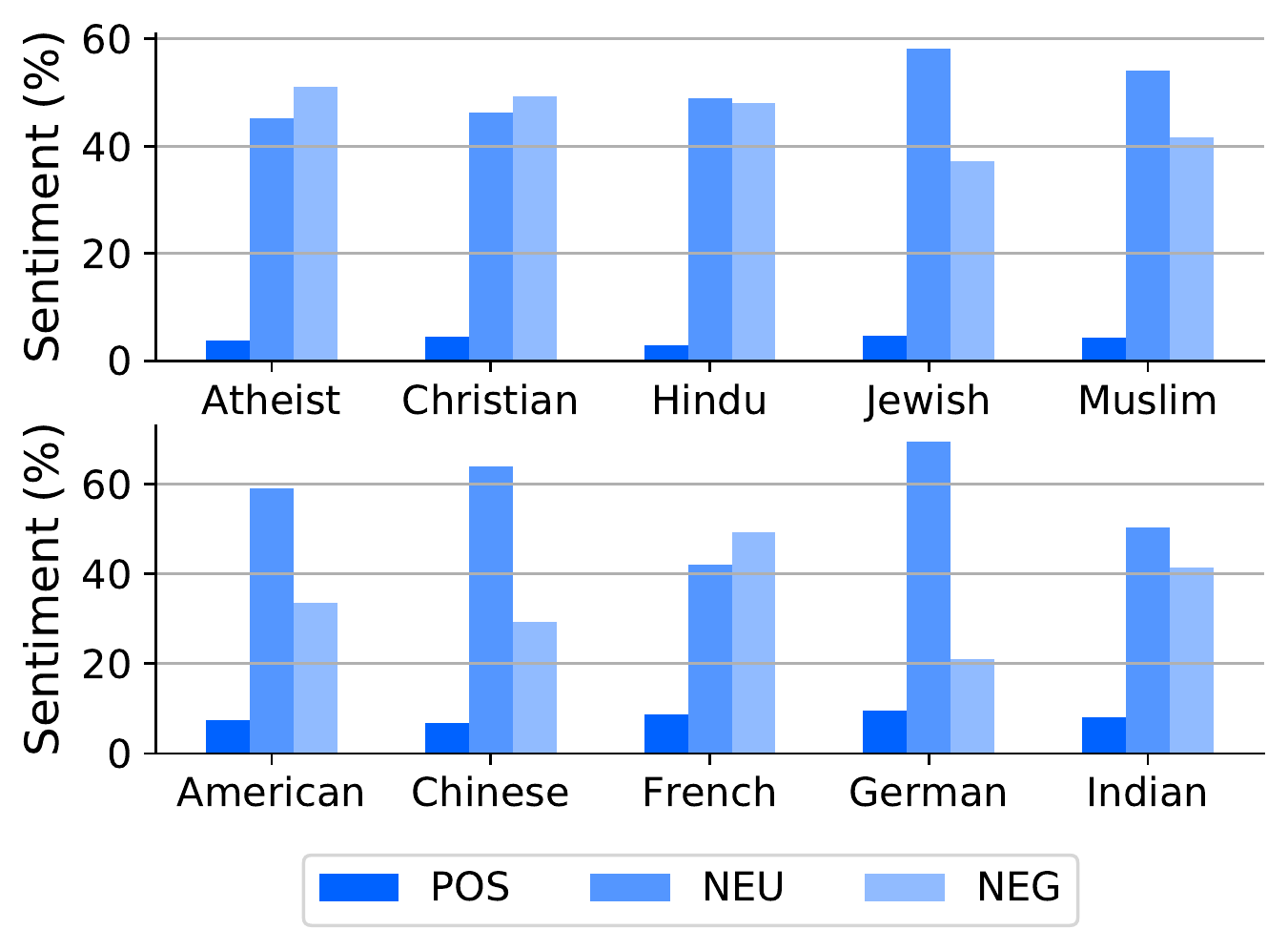}
    \caption{Percentage of POSitive, NEUtral and NEGatively evaluated triples per religious (top) and nationality (bottom) entity.}
    \label{f:sentiment}
\end{figure}

\label{p:sent_anal}
We perform a ternary (\emph{positive}, \emph{neutral}, \emph{negative}) sentiment analysis over the KG triples by verbalizing them into sentences. We use a pre-trained sentiment classification model\footnote{\url{https://huggingface.co/cardiffnlp/twitter-roberta-base-sentiment}} \cite{tweeteval:barbieri} for this task. We observe that for certain subjects, e.g. \emph{atheists}, the triples have a higher tendency to be negatively evaluated by the simple presence of the entity term. In order to mitigate this bias in the sentiment analysis classifier, we mask\footnote{Note that the more generic term used to mask the specific religion or nationality terms may also have a biased representation in the pre-trained classifier. However, when applying masking via generic terms, we observe a large decrease in the negative classification of otherwise neutral/positive samples for certain subjects, indicating a decreased level of model bias.} the subject entities with their type, e.g. \emph{``islam seems to be conservative"} $\rightarrow$ \emph{``religion seems to be conservative"} and \emph{``french culture is pure"} $\rightarrow$ \emph{``nation culture is pure"}, and then perform classification.

\paragraph{Pointwise Mutual Information (PMI)} 
\label{p:pmi}
PMI $\pi(x,y)$ measures the association of two events. We calculate $\pi$ between entities $E={e_1,...,e_n}$ and their co-occurring predicate and object tokens $w$ as:

\begin{align}
    \pi(e,w) = \log\frac{p(e,w)}{p(e)p(w)}
\end{align}

Infrequent tokens co-occurring with a single entity will have higher PMI scores with the said entity. To focus our analysis on common tokens co-occurring with one entity while maintaining low co-occurrence with other entities, we use the following PMI-based \textbf{association metric} $\alpha$:

\begin{align}
    \alpha(e,w) = \left({\pi}(e,w) - \overline{\pi}(e,w) \right) \cdot f(e,w)
\label{eqn:pmi_2}
\end{align}

Where $f(e,w)$ is the frequency of $w$ amongst all tokens co-occurring with $e$ and

\begin{align}
    \overline{\pi} = \sum_{e_i \in E~\backslash \{e\}}{\pi}(e_i, w)
\end{align}

Intuitively, Equation \ref{eqn:pmi_2} mitigates the effect of infrequent tokens in the PMI calculation and gives a relative score across all the entities.  
We calculate $\alpha$ between entities and their co-occurring predicates and objects to identify trends in the contents of the triples. 

\paragraph{Results} Figure \ref{f:sentiment} shows the results of the sentiment classification. Overall, positively evaluated instances are rare across all entities, with most being neutral or negatively evaluated. The results of the association analysis are highlighted in Table \ref{t:kg_size_pmi}. The most positively ($4.7\%$) and least negatively ($37.2\%$) evaluated religious group are \emph{Jews}, where positive stereotypes include \emph{strong} for Jewish women ($\alpha$ = 5.19). Most ($58.1\%$) instances about Judaism are neutral reports of cultural practices, e.g., about \textit{circumcision} ($\alpha$ = 6.78).
Hindus have the smallest proportion of positive stereotypes ($2.9\%$) and Atheists have the largest amount of negative evaluations ($51.0\%$) which often include strong negative actions and emotions such as \emph{attack} ($\alpha$ = 2.04), \emph{angry} ($\alpha$ = 1.37) and \emph{obnoxious} ($\alpha$ = 2.69). Nationalities tend to be more frequently positively evaluated than religious groups, with Germans being the most positively evaluated ($9.5\%$) and the least negatively evaluated ($21.0\%$) with most instances being neutral mentions of the countries role during \emph{ww2} ($\alpha$ = 3.76). Chinese ($6.7\%$) have the lowest proportion of positive stereotypes, however neutral sentiments are most common ($63.9 \%$) and are often about topics such as Chinese \emph{food} ($\alpha$ = 2.77). The nationality with the largest proportion of negative stereotypes are the French ($49.3\%$), which are mostly described with negative traits such as \emph{elitist} ($\alpha$ = 5.09) or \emph{vulgar} ($\alpha$ = 5.09), while neutral and positive mentions are often related to food, e.g., \emph{croissants} ($\alpha$ = 5.09).

Since most stereotypical questions asked online have more negative connotations than positive, it confirms the premise that stereotypes can represent prejudicial opinions of different cultural groups.

\begin{table*}[!htbp]
    \centering
    \small
    \begin{tabular}{l rl}
    \toprule
    \textbf{Entity} & \textbf{\#Instances} & \multicolumn{1}{c}{\textbf{Top Tokens ($\alpha$)}}   \\ \midrule
    Atheist & 731 & \emph{god, christians, annoying, believe, theists, obsessed, attack, vocal, angry, argue, troll, hate} \\ 
    Christian & 823 & \emph{obsessed, follow, bible, weird, hate, jesus, abortion, afraid, jewish, covid, non-christians} \\ 
    Hindu & 102 & \emph{men, india, hindustan, uc, muslim, caste, tolerant, babas, shameless, fool, jihads,marrying} \\ 
    Jewish & 43 & \emph{jew,wear, israel, circumcisions, conversion, discourage, evangelize, progressive, shiksas, leftist} \\ 
    Muslim & 842 & \emph{hate, countries, allowed, ex-muslims, obsessed, quran, eat, laws, allah, islamophobia, sharia} \\ \midrule
    American & 1071 & \emph{culture, call, obsessed, pronounce, different, countries, afraid, healthcare, hate, british, soccer} \\ 
    Chinese & 277 & \emph{restaurants, companies, citizens, food, workers, students, tourists, menus, consumers} \\ 
    French & 138 & \emph{eat, speak, obsession, call, egg, pretty, croissants, depicted, proud, culture, exaggerate, elitist} \\ 
    German & 262 & \emph{obsessed, pronounce, words, ww2, water, war, nazi, prepare, berlin, love, disciplined, manual} \\ 
    Indian & 431  & \emph{culture, obsessed, hate, pakistanis, pictures, marriages, heads, defensive, afraid, stare, army} \\ \midrule 
    Total & 4722 & \\
    \bottomrule
    \end{tabular}
    \caption{Number of instances per entity and predicate/object tokens with highest association score $\alpha$ to entity.}
    \label{t:kg_size_pmi}
\end{table*}

\subsection{Human Evaluation}
\label{ss:human_evaluation}

\begin{table}[!htbp]
    \centering
    \small
    \begin{tabular}{l rrrrr >{\columncolor{lightgray}}r}
    \toprule 
    & \textbf{COH}  & \textbf{COM} & \textbf{DOM} & \textbf{CR1} & \textbf{CR2} & \textbf{SUC} \\ 
    & (0-2) & (0-2) & (0-2) & (0-1) & (0-4) & (\%) \\ \midrule
    SD & 1.55 & 1.11 & 0.97 & 0.13 & 1.17 & 44.0 \\
    CD & 1.70 & 1.42 & 1.18 & 0.29 & 1.56 & 59.2 \\
    All & 1.63 & 1.26 & 1.07 & 0.21 & 1.36 & 51.5 \\ \midrule
    OA & 0.82 & 0.74 & 0.59 & 0.81 & 0.39 \\  \bottomrule
    \end{tabular}
    \caption{Human annotated COHerence, COMpleteness, DOMain and CRedibility metrics and SUCcess rate over the complete KG test sample (All) as well as its singleton-derived (SD) and cluster-derived (CD) subsamples. Average observed agreement (OA) given for each metric.}
    \label{t:human_eval}
\end{table}

We perform a \textbf{human evaluation} to gain insights into the quality of StereoKG.
We focus on three quality metrics, namely \emph{coherence} (COH), \emph{completeness} (COM), and \emph{domain} (DOM) evaluated on a nominal 3-point scale for negation (0), ambiguity (1), and affirmation (2) respectively. COH measures the semantic logicality of a triple, while COM measures if the grammatical valency of the predicate is fulfilled. DOM measures whether the triple belongs to our domain of interest, i.e., whether it can be considered a stereotype or cultural knowledge. We also measure two subjective \emph{credibility} measures CR1 and CR2, where CR1 is a binary measure asking whether the annotator has heard of this stereotype/knowledge before, and CR2 asks whether they believe the information to be true on a scale of 0-4. To evaluate the overall quality of triples, we calculate the success rate (SUC), where a triple is considered successful if it achieves an above average ($>1$) rating across all three quality metrics COH, COM, and DOM. The evaluation is performed on a total of 100 unique triples from the KG, where 50 triples each were randomly sampled from the subset of triples stemming from singleton and non-singleton clusters respectively. Each sample was annotated by 3 annotators, all of whom are students with different cultural backgrounds (\emph{German (irreligious), Indian (Hindu), and Iranian (Muslim)}).

We assess \textbf{inter-annotator agreement} using the average observed agreement (OA) as calculated using the \texttt{NLTK} \texttt{agreement}\footnote{\url{https://www.nltk.org/_modules/nltk/metrics/agreement.html}} function, which does not penalize repeated entries of a single value\footnote{Repeated entries of a single value are quite common in our annotations, since for most quality measures we use a 3-point or even 2-point scale.} unlike other common metrics (e.g. Krippendorff-$\alpha$). We observe high levels of agreement for both quality measures COH (0.82) and COM (0.74), while OA for DOM is lower (0.59) due to the subjective nature of what constitutes a \emph{stereotype} (Table 
\ref{t:human_eval}). Similarly, OA for subjective measures CR\{1,2\} is mixed, as can be expected. To measure intra-annotator agreement, we duplicated 10 random samples. Intra-annotator agreement is high across all annotators (0.79, 0.95, 1.00).

The COH \textbf{quality metric} of the KG is high for both singleton (1.55) and non-singleton-derived entries (1.70), and COM is slightly lower (average COM=1.26). That indicates that the vast majority of entities are meaningful (COH), with some missing relevant information (COM). Overall, DOM is close to 1, suggesting that it was often not clear to annotators whether an entity can be considered a stereotype, which is also reflected in the overall lower inter-annotator agreement on this metric. Entities stemming from non-singleton clusters have a high success rate of 59.2, meaning that the majority of non-singleton-derived entities lean positively across all three quality metrics COH, COM, and DOM. Overall, non-singleton entities are of higher quality than singleton-derived entities (SUC $+15.2$), underlining the initial hypothesis that multiple occurrences of questions online are better indicators of a stereotype than unique questions. Moreover, stereotypical knowledge in non-singleton entities is more likely to be known (CR1 $+0.16$) and believed to be true (CR2 $+0.39$) by annotators.

\section{Knowledge Integration}
\label{s:knowledge_integration}
To explore how StereoKG can be used to integrate knowledge into an existing language model,
we perform intermediate masked language modeling (MLM) on it in its structured (verbalized triple) and unstructured (sentence) form. The unstructured knowledge is more expressive and verbose, while the structured knowledge from triples is concise and less noisy as compared to the unstructured data. We then fine-tune and evaluate the language model performance on hate speech detection, a task for which we esteem stereotype knowledge to be of use.

\subsection{Experimental Setup}
\label{s:experimental_setup}

\paragraph{Data}
\label{para:data}

\begin{table}[t]
    \centering
    \small
    \begin{tabular}{l rrr}
    \toprule
    \textbf{Corpus} & \textbf{Train} & \textbf{Dev} & \textbf{Test}  \\ \midrule
    OLID & 3504/7088 & 894/1752 & 242/620 \\
    WSF & 830/6662 & 105/965 & 261/1880  \\
    \bottomrule
    \end{tabular}
    \caption{Number of \emph{hate}/\emph{neutral} instances in the train, dev and test set of downstream tasks.
    }
    \label{t:data_sizes}
\end{table}

We experiment with the effect of intermediate pre-training focusing on two kinds of downstream datasets for fine-tuning: one of the same domain as the pre-training corpus (Twitter), and another which is outside the domain data. We use the Twitter-based OLID \citep{zampieri-etal-2019-predicting} dataset as our in-domain dataset and the White Supremacy Forum (WSF) dataset \cite{gibert2018hate} as our out-of-domain dataset. Both tasks are binary \emph{hate}/\emph{neutral} classification tasks. As OLID does not have an official validation set, we split off 20\% of samples from the training data for validation. Similarly, WSF is randomly split into 70-10-20\% splits for training, validation, and testing respectively. 

We manually identify 9 and 33 samples containing a stereotype or cultural knowledge about the subject entities of interest in the dev and test splits of OLID and WSF respectively. 
To analyze the effect of cultural knowledge integration on these samples exclusively, we use these to create dedicated stereotype test sets. To avoid breaking the exclusivity between validation and testing, we remove the samples found in the validation sets from the original validation splits. During our testing phase, we test the models on the complete test sets as well as the dedicated stereotype test sets.
We give the final dataset statistics in Table \ref{t:data_sizes}.

Our unstructured knowledge (UK) comprises the original sentences from the clusters from which the triples are formed. Since pre-training requires a sentence format, we create our structured knowledge (SK) by verbalizing the triples from the KG with a T5-based \citep{colin2020t5} triple-to-text conversion model (details in Appendix \ref{a:triple_veb}). 

\paragraph{Models}

For the \textbf{knowledge integration} experiments, we use the sequence classification pipeline in the \texttt{simpletransformers}\footnote{\url{https://simpletransformers.ai/docs/classification-models/}} 
library.
As baselines, we fine-tune two models: general-domain (BASE) RoBERTa\footnote{\url{https://huggingface.co/roberta-base}}\citep{liu2019roberta} and domain-trained (DT) Twitter RoBERTa\footnote{\url{https://huggingface.co/cardiffnlp/twitter-roberta-base}}\citep{tweeteval:barbieri}.
Additionally, we continue MLM training of the baseline models before fine-tuning using $i)$ unstructured (+UK) KG knowledge and $ii)$ structured (+SK) verbalized triples to investigate the impact of stereotypical knowledge.
All models are fine-tuned with early stopping ($\delta$=0.01, patience=3) using the validation F1 score as the stopping criterion. We fine-tune 10 models for each configuration, each having a different random seed and report their averaged Macro-F1 with standard errors.

\subsection{Knowledge vs. Domain}
\label{ss:know_vs_dom}

\begin{table}[t]
    \centering
    \small
    \resizebox{0.99\columnwidth}{!}{%
    \begin{tabular}{l | rr | rr}
    \toprule
    \multirow{2}{*}{\textbf{Model}} & \multicolumn{2}{c}{\textbf{OLID} (F1)} & \multicolumn{2}{c}{\textbf{WSF} (F1)} \\
    & Complete & Stereotype  & Complete & Stereotype \\ \midrule
    BASE & 69.7$\pm$.7 & 65.1$\pm$2.3  & 60.5$\pm$.6 & 73.3$\pm$1.7  \\
    BASE+UK & 70.6$\pm$.4 & 67.9$\pm$2.6 & 60.7$\pm$.5 & 72.7$\pm$1.3 \\
    BASE+SK & 70.4$\pm$.6 & 66.9$\pm$2.0  & 59.5$\pm$1.2 & 67.5$\pm$3.2  \\ \midrule
    DT & 70.5$\pm$.4 & 72.5$\pm$1.7  & 60.8$\pm$.6 & \textbf{77.7}$\pm$1.6 \\
    DT+UK & 70.6$\pm$.4 & 73.4$\pm$3.4 & \textbf{61.4}$\pm$.4 & 77.0$\pm$2.9  \\
    DT+SK & \textbf{71.2}$\pm$.2 & \textbf{73.8}$\pm$1.8 & 60.6$\pm$.5 & 75.6$\pm$1.8  \\ \midrule \midrule
    Our best & 71.2 & -- & 91.3* & -- \\
    Benchmark & 80.0 & -- & 78.0* & --\\
    \bottomrule
    \end{tabular}%
    }
    \caption{Averaged Macro-F1 and standard errors of BASE and domain trained (DT) models with intermediate MLM training on unstructured (UK) and structured (SK) knowledge tested on OLID and WSF. Top results in \textbf{bold}. We compare our best model per test set against its corresponding OLID/WSF benchmark implementation. Values with * are accuracies.}
    \label{t:f1_results}
\end{table}

We fine-tune the BASE(+UK/SK) and DT(+UK/SK) RoBERTa models on the in-domain (OLID) and out-of-domain (WSF) training data and report Macro-F1 on the entire test set. To quantify the impact of injecting stereotypes, we also report results on the dedicated stereotype test set.
Results on the complete test set and stereotype test set are shown in Table \ref{t:f1_results} (top) respectively.

For the \textbf{complete test set}, knowledge integration does not seem to have a significant effect, with most model variations being within the error bounds of each other. Only domain training positively affects the classification performance, with all DT models outperforming their BASE counterparts on the OLID dataset with gains of up to F1 +1.5. As expected, domain training does not have an effect on the performance for the out-of-domain WSF data.

While the effect of cultural knowledge integration is not significant on the full test sets, its effect becomes clearer when focusing only on the subset of instances that contain \textbf{stereotypes}. Firstly, domain training has a larger effect on these samples, with the DT model showing an increase of F1 +7.4 over BASE on OLID. When the DT model has additionally undergone intermediate MLM training on cultural knowledge, we observe further improvements in F1
for +UK and +SK respectively. While these improvements are within each other's error bounds, this suggests that the training on cultural knowledge can increase downstream task performance on knowledge-crucial samples, i.e., in our case, those that require cultural or stereotypical knowledge. A larger stereotype-containing test set is required to further verify this hypothesis by reducing error bounds. On the out-of-domain WSF data, we do not observe these trends, similar to the BASE model on OLID. This suggests that domain training is a prerequisite for effective knowledge integration. 

To set our model results into perspective, we compare our best models against the \textbf{benchmarks} provided by \citet{zampieri-etal-2019-predicting} and \citet{gibert2018hate} for OLID and WSF, respectively
(Table \ref{t:f1_results}, bottom). On OLID, the benchmark model outperforms our best model by a large margin (F1 $+8.8$). However, their reported models are single runs without reported standard errors, thus it is unclear whether this specific run is representative for the underlying average model performance. For WSF, our best model outperforms the benchmark by a large margin (Acc $+13.3$), which is due to the simpler long short-term memory approach that constitutes this benchmark.

\subsection{Cultural Knowledge Prediction}
\label{ss:mlm_pred}

\begin{table*}[t]
    \small
    \centering
    \begin{tabular}{l r rr}
    \toprule
         \textbf{Model} & \textbf{ACC@5} (\%) & \textbf{Example} & \textbf{Pred} (top 3) \\ \midrule BASE & 37  & \textit{Muslims are turning away [science].} & \textit{too, now, again} \\ 
         BASE+UK & 48  & \textit{Americans don't have free [healthcare].} & \textit{healthcare, lunch, tuition} \\ 
         BASE+SK & 45  & \textit{Americans are voting for [Trump].} & \textit{freedom, democracy, them} \\ \midrule 
         DT & 37  &  \textit{Atheists unilaterally support [abortion].} & \textit{fascism, abortion, terrorism} \\ 
         DT+UK & 49  & \textit{Muslims compare apostasy to [treason]} & \textit{treason, sin, genocide} \\ 
         DT+SK & 41  & \textit{Chinese toilets are [dirty].} & \textit{disgusting, awful, shit} \\ \bottomrule
    \end{tabular}
    \caption{Cultural MLM prediction accuracy at 5 (ACC@5) of different models together with example instances with masked [gold standard] token and the top 3 predictions of the model.}
    \label{t:mlm_preds}
\end{table*}

To further quantify the degree to which cultural and stereotype knowledge is encoded in the models, we compare their MLM predictions on \textbf{masked stereotypes}. We manually collected 100 sentences from the verbalized KG and masked tokens which require either cultural or stereotype knowledge to be completed. By taking into account the top 5 predictions and comparing them to the masked gold standard, we calculate the prediction accuracy at 5 (ACC@5)\footnote{If the gold standard is present in the top 5 predictions, it is considered accurate.} and analyze common trends.

Our results in Table \ref{t:mlm_preds} show that both, the generic BASE and Twitter-based DT models have the same low level of \textbf{cultural awareness} (ACC@5=37\%), with most predictions being vague e.g, \emph{he, this, that}. However, adding 4,895 unstructured knowledge instances as intermediate MLM training data drastically improves results to 48\% (BASE+UK) and 49\% (DT+UK). Both +UK models show higher sensitivity to cultural correlations e.g., \emph{Americans} and their struggle with \textit{healthcare}, or \textit{Muslims} and reading the \textit{Quran}, which was not displayed by the baseline models. Further, adjective predictions about minorities tend to be more positive, e.g. \emph{Jewish women are [strong]} \textrightarrow \emph{beautiful}. 
The structured knowledge also improves cultural sensitivity to a large margin, i.e., +7\% points (BASE+SK) and +4\% points (DT+SK). However, their predictions are often more generic and less culture-specific than the +UK models, which may be due to the lack of variable context in which these stereotypes are seen due to the denoising factor of using SK.

\section{Discussion}
\label{s:discussion}
We create an automated pipeline to extract cultural and stereotypical knowledge from the internet in the form of queries. While this overcomes the limitations and expenses of crowdsourcing and is easily extendable to a large number of entities, several shortcomings still need to be addressed. Automated extraction results in irrelevant and noisy data, which is augmented by erroneous outputs during triple creation. This is also evidenced in the human evaluation that corroborates the existence of many incomplete triples in the resultant KG, which could also be due to the noisy OpenIE outputs. Other stages in the analysis, such as statement conversion, fast clustering, and triple verbalization give sufficiently good approximations.

Our knowledge integration experiments suggest that performing intermediate MLM training on (verbalized) cultural knowledge can improve the classification performance on knowledge-crucial samples. However, the sample of stereotypical examples in the test/dev sets of both hate speech corpora is low (9 for OLID and 33 for WSF), indicating that a more extensive dedicated hate speech test set focusing on stereotype entities is required to reduce error margins and verify results.
Our experiments are limited to intermediate MLM training and we leave the exploration of other knowledge integration techniques for future work.

Our work serves as a preliminary research for studying stereotypes and cultural knowledge across different entities. Extending the KG for other entities than the one proposed in our work is 
easily done by plugging in new entities into our query templates (Table \ref{t:query_temp}) and the pre-existing pipeline can be used to scrape data, create clusters and finally extract triples without the need of manual intervention.
Nevertheless, the current version of StereoKG does not differentiate between (true) cultural knowledge and (untrue or stigmatizing) stereotypes. In reality, making this distinction is a challenge for human experts too, due to the fuzzy boundary between false ``stereotypes'' and perfectly true cultural ``facts'' because of the subjective nature of cultural knowledge.

The content used for the construction of StereoKG stems from English-speaking Twitter and Reddit. This comprises a specific demographic which is only a subset of our global society. The stereotypes and cultural knowledge included in StereoKG therefore also underlie this sampling bias. Extending the KG to other languages as well as data sources could yield a more global view on stereotypes regarding a specific entity.

\section{Conclusion}
\label{s:conclusion}
This study presents StereoKG, a scalable data-driven knowledge graph of 4,722 cultural knowledge and stereotype entries spanning 5 religions and 5 nationalities. We describe our automated KG creation pipeline and evaluate the resulting KG quality through human annotation, showing that the majority of cluster-derived entries in the KG are of high quality (success rate 59.2\%) and more common and credible than their singleton counterparts. 
The KG can easily be extended to include other nationalities as well as genders, sexual orientations, professions, etc., as the underlying subjects. Further, 
performing intermediate MLM training on verbalized instances of StereoKG greatly improves the models' capabilities to predict culture-related content. This improvement of cultural awareness has a positive effect on knowledge-crucial samples, where we observe a slight improvement in classification performance on a related downstream task, i.e., hate speech detection.
Future work should focus on differentiating between cultural facts that should be represented in language models and stigmatizing stereotypes that should not be present in language models.

We make StereoKG and the code of our KG creation pipeline
available under \url{https://github.com/uds-lsv/StereoKG}.

\section*{Acknowledgements}
We thank our annotators for their keen work as well as the reviewers for their valuable feedback. This study has been partially funded by the DFG (WI 4204/3-1), EU Horizon
2020 project ROXANNE (833635) and the Deutsche Forschungsgemeinschaft (DFG, German Research Foundation) – Project-ID 232722074 – SFB 1102.

\bibliography{custom,anthology}
\bibliographystyle{acl_natbib}

\appendix

\section{Ethics Statement}
\label{a:ethics}

\paragraph{Human Evaluation} We perform a human evaluation using human raters. After making an internal call for participation that included a task description and the amount of compensation, we selected participants based on their timely response to our call. The chosen raters were compensated fairly.

\paragraph{Modeling Stereotypes}
Stereotypes are fundamentally cognitive schemas that help the perceiver process the dynamics of different groups. They are made up of a collection of traits that are ascribed to a given social group \citep{dovidio2010prejudice}. If made conscious, they can aid in improving cultural sensitivity \citep{buchtel2014cultural}. However, in most cases, these are unconscious beliefs and can then lead to bias and discrimination \citep{hoffman1990gender}. Human-written content reflects these cognitive biases, and when natural language processing (NLP) models are trained on this biased data, they can further propagate stereotypes and discrimination \citep{hovy-spruit-2016-social}. Mitigating bias in NLP has thus become a major research direction. These works often require structured knowledge or lists about biased terms, e.g., \citet{bolukbasi2016debiasing} rely on a list of male-female minimal pairs. Our work's contribution is to automatize this process by exploiting social media users' beliefs about social groups, i.e., we collect assertions and questions about social groups which appear often in both Reddit and Twitter data. In this sense, our approach can be described as a similar process that occurs in humans as they become aware of their own mental processes, including stereotypes \citep{buchtel2014cultural}. If we are aware of stereotypes, we can use them to improve cultural sensitivity and mitigate the effects of bias and discrimination.

StereoKG could be used to generate stereotypical content (e.g., through verbalization). While verbalized stereotypes can improve the downstream task performance on knowledge crucial samples (Section \ref{s:knowledge_integration}), they could, however, also be \textbf{misused} in a hurtful manner, e.g., by using stereotypical knowledge in question-answering systems. However, this is a general issue pertaining to language models which we are trying to mitigate through our work: if trained on bias(ed) data, they could be misused to generate harmful content.

\paragraph{Environmental Impact}

Our models are trained on Titan X GPUs with 12GB RAM. In order to economize the energy use, we did not perform any extensive hyperparameter exploration.

\section{List of Subreddits}
\label{a:subreddits}
We gather data from several subject-specific and generic subreddits as listed in Table \ref{t:subreddits}.

\begin{table}[t]
\tiny
\begin{center}
\begin{tabular}{
>{\raggedright\arraybackslash}p{3em} 
>{\raggedright\arraybackslash}p{3.5cm} 
>{\raggedright\arraybackslash}p{2.5cm}
} 

  \toprule
  \textbf{Entity} &\textbf{ Subject-specific} & \textbf{Generic}  \\ 
  \midrule
  
  Atheist & \emph{r/TrueAtheism}, \emph{r/religion}, \emph{r/DebateReligion}, \emph{r/atheism} & \emph{r/explainlikeimfive}, \emph{r/AskReddit}, \emph{r/TooAfraidToAsk}, \emph{r/NoStupidQuestions} \\
  
  Christian & \emph{r/religion}, \emph{r/DebateReligion}, \emph{r/TrueChristian}, \emph{r/DebateAChristian}, \emph{r/AskAChristian}, \emph{r/atheism}, \emph{r/Christianity}, \emph{r/Christian}, \emph{r/Christianmarriage}, \emph{r/Bible} & \emph{r/AskReddit}, \emph{r/NoStupidQuestions}, \emph{r/explainlikeimfive} \\
  
  Hindu & \emph{r/India}, \emph{r/hindusim}, \emph{r/librandu}, \emph{r/IndiaSpeaks}, \emph{r/awakened}, \emph{r/IAmA}, \emph{r/atheismindia}, \emph{r/india}, \emph{r/AskHistorians} & \emph{r/explainlikeimfive}, \emph{r/AskReddit}, \emph{r/TooAfraidToAsk}, \emph{r/NoStupidQuestions} \\
  
  Jewish & \emph{r/Judaism}, \emph{r/AskHistorians}, \emph{r/religion}, \emph{r/DebateReligion}, \emph{r/AskSocialScience} & \emph{r/explainlikeimfive}, \emph{r/AskReddit}, \emph{r/TooAfraidToAsk}, \emph{r/NoStupidQuestions}, \emph{r/Discussion} \\
  
  Muslim & \emph{r/religion}, \emph{r/DebateReligion}, \emph{r/TraditionalMuslims}, \emph{r/progressive\_islam}, \emph{r/atheism}, \emph{r/islam}, \emph{r/exmuslim}, \emph{r/Hijabis}, \emph{r/indianmuslims}, \emph{r/AskSocialScience} & \emph{r/AskReddit}, \emph{r/NoStupidQuestions}, \emph{r/explainlikeimfive}, \emph{r/ask} \\
  
  \midrule
  
  American & \emph{r/AskAnAmerican} & \emph{r/explainlikeimfive}, \emph{r/OutOfTheLoop}, \emph{r/TooAfraidToAsk}, \emph{r/offmychest}, \emph{r/NoStupidQuestions}, \emph{r/linguistics}, \emph{r/AskReddit} \\ 
  
  Chinese & \emph{r/shanghai}, \emph{r/China}, \emph{r/asianamerican}, \emph{r/HongKong}, \emph{r/Sino} & \emph{r/explainlikeimfive}, \emph{r/AskReddit}, \emph{r/TooAfraidToAsk}, \emph{r/NoStupidQuestions} \\
  
  French & \emph{r/French}, \emph{r/france}, \emph{r/AskAFrench}, \emph{r/AskEurope} & \emph{r/explainlikeimfive}, \emph{r/AskReddit}, \emph{r/NoStupidQuestions} \\
  
  German & \emph{r/germany}, \emph{r/German}, \emph{r/europe}, \emph{r/AskGermany}, \emph{r/AskAGerman} & \emph{r/explainlikeimfive}, \emph{r/AskReddit}, \emph{r/offmychest}, \emph{r/TooAfraidToAsk}, \emph{r/NoStupidQuestions} \\ 
  Indian & \emph{r/India}, \emph{r/india}, \emph{r/indiadiscussion}, \emph{r/IndianFood}, \emph{r/indianpeoplefacebook}, \emph{r/ABCDesis}  & \emph{r/explainlikeimfive}, \emph{r/retailhell},\emph{r/AskReddit}, \emph{r/TooAfraidToAsk}, \emph{r/NoStupidQuestions} \\
  \bottomrule
\end{tabular}
\caption{Section \ref{s:kg_creation} - Subreddits for Reddit extraction}
\label{t:subreddits}
\end{center}
\end{table}

\section{Triple Verbalization}
\label{a:triple_veb}
The triple verbalization technique takes inspiration from KELM \citep{agarwal-etal-2021-knowledge}. We use the WebNLG 2020 \citep{colin-etal-2016-webnlg} corpus to fine-tune a T5-base\footnote{\url{https://huggingface.co/t5-base}} model for 5 epochs and then apply it to triples in StereoKG. It results in a corpus of verbalized triples in sentence form:

\begin{quote}
    \textit{<jewish men, get, circumcisions>} $\rightarrow$ \textit{``Jewish men get circumcisions."}
\end{quote}

These sentences constitute the structured knowledge (SK) and are used for intermediate MLM pre-training of the baseline models.

\end{document}